# On a heuristic approach to the description of consciousness as a hypercomplex system state and the possibility of machine consciousness (German edition)

Ralf Otte

Ulm University of Applied Sciences
Prittwitzstraße 10
Germany - 89075 Ulm

ralf.otte@thu.de
ralfotte.com

Ibiza, May, 30th, 2024

**Summary**

This article presents a heuristic viewpoint that shows that the inner states of consciousness experienced by every human being have a physical, but imaginary, hypercomplex basis. Based on theoretical considerations about human consciousness, it will very probably be possible to make targeted technical use of hypercomplex system states in the future as a result of mathematical investigations.

The hypothesis of the existence of hypercomplex system states on technical machines is already supported by the performance of highly complex AI systems. However, this has yet to be proven. In particular, there is a lack of experimental data that distinguishes such systems from other systems, which is why this question will be addressed in subsequent articles. This paper shows a mathematical possibility that hypercomplex system states on machines could actually be generated and used in a targeted manner. In the literature, hypercomplex system states are already referred to as machine consciousness, it could become really possible.

### Einführung und Vorbemerkungen

In diesem Beitrag wird als Postulat vorausgesetzt, dass das menschliche Bewusstsein eine real existierende ontologische Existenz besitzt, die durch die Naturwissenschaften untersucht werden sollte und mit wissenschaftlichen Methoden auch untersucht werden kann. Bei den in diesem Beitrag vorgeschlagenen Untersuchungen geht es jedoch nicht um alle möglichen Aspekte von Bewusstsein, die durch Neurologen, Psychiater, Soziologen und andere bereits untersucht werden, sondern ausschließlich um die Beschreibung rein physikalischer Aspekte des Bewusstseins, da Bewusstsein einen physikalischen Hintergrund besitzen muss, wenn es objektiv existiert. In diesem Aufsatz wird deshalb der Versuch unternommen, Bewusstseinszustände mathematisch als sogenannte *hyperkomplexe Systemzustände* zu beschreiben, um sie einer objektiven naturwissenschaftlichen Behandlung zuzuführen.

Manche Leser könnten einwenden, dass Bewusstsein nicht objektiv, sondern nur subjektiv existiert. Aber die Gesamtheit aller „subjektiven Bewusstseine" von Menschen existiert zwangsläufig objektiv, zumindest in dem Sinne, dass beispielsweise die „Bewusstseine" der Leserinnen und Leser nicht vom subjektiven Bewusstsein des Autors abhängen, und damit als objektiv existent, d.h., unabhängig vom Autor, angenommen werden müssen. Aus Sicht des Autors ist daher eine naturwissenschaftliche Beschreibung des Bewusstseins notwendig. Glaubensvorstellungen oder auch ein neuer Beitrag zu den Jahrzehnte-währenden philosophischen Diskussionen zum Bewusstsein sind ausdrücklich nicht Ziel dieser Arbeit, da, nach Auffassung des Autors, man ab einer gewissen Beschreibungstiefe mit dem philosophischen Diskurs nicht weiterkommt. In diesem Artikel wird daher ein einfacher ingenieurmäßiger Standpunkt vertreten: Wenn Bewusstsein existiert, muss es sich *irgendwie* physikalisch bemerkbar machen. Punktum. Es geht daher um die naturwissenschaftliche Analyse von physikalischen Eigenschaften von Bewusstsein und - in der Folge - um erste Möglichkeiten zu deren Umsetzung auf technischen Systemen.

Bewusstsein wird aus den Naturwissenschaften oft ausgeklammert, da bisher kein Verfahren existiert, um es objektiv zu vermessen. Doch Nichtmessbarkeit einer Eigenschaft sollte aus Sicht des Autors kein Hinderungsgrund für physikalische Analysen sein. Selbstverständlich ist das ein strittiger Standpunkt. Geradezu berühmt wurde ein diesbezüglicher Streit zwischen Einstein vs. Bohr und Heisenberg, ob die Physik die Wirklichkeit *wie sie ist* beschreiben sollte (Einstein) oder nur das beschreiben kann, *was messbar ist* (Bohr, Heisenberg). Jeder dieser Ansichten hat Vor- und Nachteile. Bei der Untersuchung von Bewusstsein sollte seine Nichtmessbarkeit jedoch nicht zu seinem Ausschluss aus mathematischen und physikalischen Analysen führen.

**ÜBER DIE NICHTMESSBARKEIT VON BEWUSSTSEINSPHÄNOMENEN**

Jeder Mensch weiß um seine inneren subjektiven Bewusstseinsinhalte, zum Beispiel um seine subjektiven Bilder, Farben oder Geräusche im Kopf. Aus naturwissenschaftlicher Sicht wird es also darum gehen, die Grundlagen solcher Speicherungen zu ermitteln, denn jedes Speichern von Informationen muss physikalische Grundlagen besitzen. Es ist unstrittig, dass die subjektiven Erlebnisinhalte eines Menschen nicht durch äußere Messungen am Gehirn, wie fMRT oder EEG, ermittelbar sind (höchstens deren „komplementäre" neuronale Korrelate). Aus der Nichtmessbarkeit von Phänomenen deren Existenz abzustreiten entbehrt jedoch jeder Kenntnis der Geschichte der Naturwissenschaften. So wie seinerzeit unsichtbare elektrische oder magnetische Felder nicht messbar waren, so könnten auch Bewusstseinszustände mit der vorliegenden Messtechnik nicht messbar sein.

Allerdings soll hier nicht der Eindruck vermittelt werden, dass Bewusstseinsinhalte bald messtechnisch erfasst werden können. Im Gegenteil, es wird im Folgenden gezeigt werden, dass Bewusstseinsphänomene sich auch zukünftig jeder direkten physikalischen Messung entziehen werden. Doch selbst das darf nicht dazu führen, eine wissenschaftliche Analyse zu unterlassen, wie bestimmte Probleme der Quantenphysik – zum Beispiel das unmögliche Beobachten von Bahnkurven der Elektronen im Atom – verdeutlicht haben.

Um dem Ergebnis des Aufsatzes vorzugreifen: Selbst dann, wenn Bewusstseinszustände nicht direkt messbar sind, so wird aufgezeigt werden, dass sich (bestimmte) Wirkungen von Bewusstseinszuständen sehr gut messen lassen und sich darüber das Vorhandensein von Bewusstsein auf einem System zumindest indirekt überprüfen lässt. Benutzt man rein physikalische Messapparaturen, so wird man selbstverständlich nur physikalische Wirkungen von Bewusstsein messen können. Biologische oder gar soziale Wirkungen von Bewusstsein können mit physikalischen Apparaturen naturgemäß nicht erfasst werden. Das wird aber auch keiner erwarten.

Die Frage, die sich zwingend stellt, ist, wieso es weltweit bis heute nicht gelungen ist, die subjektiven (mentalen) Bewusstseinszustände von Menschen zu messen, wenn sie doch wie hier postuliert, physikalisch objektiv vorhanden sind. Der Grund liegt aus Sicht des Autors an der Art des Energieinhaltes der Bewusstseinszustände. Mit heutiger Technik lassen sich nur Zustände und Felder messen, die einen *realen* Energieinhalt besitzen; ob der Energieinhalt durch elektrische, magnetische oder gravitative Felder hervorgerufen wird, ist dabei nicht entscheidend. Doch Bewusstseinszustände besitzen wahrscheinlich keine solche Energieform. Die „geistigen" Prozesse, die neben den neuronalen Prozessen im Gehirn die subjektiven Bewusstseinsphänomene ermöglichen, haben höchstwahrscheinlich überhaupt keinen real existierenden Energieinhalt. Genau deshalb sind sie eben nicht messbar, obwohl sie existieren. Das soll im Weiteren verdeutlicht werden.

Bisher sind bei naturwissenschaftlichen Prozessen nur Energieformen bekannt, die man mit reellen Zahlen formalisieren kann. Normalerweise erhält man für natürliche Prozesse jedweder Art positive reelle Energie(eigen)werte. Paul Dirac fand 1928 bei seinen theoretischen Überlegungen jedoch erste Hinweise auf negative reelle Energien, die später als bestimmte Form der Materie gedeutet wurden. Jedoch kann man bereits durch rein formale Überlegungen den Energiebegriff erweitern. Man kann, wie in der Mathematik üblich, senkrecht zur reellen Zahlengeraden eine weitere Gerade einführen, die die sogenannten imaginären Zahlen *i* abbildet. Dadurch entsteht eine Ebene, die sogenannte *Gaußsche Zahlenebene*, die komplexe Zahlen der Form $z=a+bi$ als Punkte (a,b) in dieser Ebene darstellen kann. (Der reelle Teil „*a*" wird auf der reellen Achse und der imaginäre Teil „*bi*" auf der imaginären Achse abgetragen.) Komplexe Zahlen werden in den Naturwissenschaften vielseitig genutzt. Seit Jahren werden in der Physik bereits Entitäten mit imaginären oder komplexen Massen diskutiert, zum Beispiel Tachyonen [1].

Es wäre zumindest denkbar, dass bestimmte Prozesse der Natur auch imaginäre oder komplexe Energiezustände besitzen, auch wenn diese bis heute noch nicht beobachtet wurden. Es wird nun explizit angenommen, dass Bewusstseinsprozesse solche ungewöhnlichen Prozesse sind. Allerdings wird eine mathematische Analyse zeigen, dass die komplexen Zahlen nicht ausreichen, um die Besonderheiten von Bewusstseinsphänomen zu formalisieren.

## MATHEMATISCHE FORMULIERUNG VON HYPERKOMPLEXEN SYSTEMZUSTÄNDEN

Mathematische Modelle werden seitjeher benutzt, um physikalische Phänomene zu vereinfachen und zu verstehen. Entziehen sich die zu beschreibenden Prozesse prinzipiell einer Modellierung, kann es daran liegen, dass mit einer ungeeigneten Algebra gearbeitet wird. Hier ein Trivialbeispiel: Will man beispielsweise betriebswirtschaftliche Phänomene von Schulden mathematisch korrekt modellieren, benötigt man Zahlen, die kleiner als null werden können. Die Algebra der natürlichen Zahlen (0, 1, 2, 3, …) wäre hier die falsche Wahl, denn die Algebra bestimmt die Rechenregeln (+-*/), und manche Rechenoperationen sind in manchen Algebren nicht erlaubt. Beispielsweise ist es in der Algebra der natürlichen Zahlen nicht möglich, die Aufgabe *„5–12"* zu lösen, denn negative Zahlen sind dort nicht existent. Auch gibt es bei natürlichen Zahlen keine Brüche, wie *„7/15"*. Deshalb wird auch seit Jahrhunderten die Berechnung von Ackerflächen nicht mit natürlichen Zahlen durchgeführt, sondern mit der (richtig ausgewählten) Algebra der rationalen bzw. reellen Zahlen. Insbesondere mit reellen Zahlen konnten und können fast alle Probleme in Gesellschaft und Wissenschaft mathematisch korrekt bearbeitet werden. Dennoch benötigte man bereits im ausgehenden Mittelalter neue Zahlen, namentlich die oben eingeführten *komplexen Zahlen*, um bestimmte mathematische Spezialaufgaben zu lösen.

Um mit Zahlen der Form *a+bi* zu rechnen, wurde die komplexe Algebra *C* geschaffen. Berechnungen mit der Einheit *i* werden durch die Multiplikationstabelle festgelegt, Abbildung 1, linke Grafik, linker oberer (hellblauer) Quadrant: *1·1=1, 1·i=i, i·1=i* und *i·i=-1*. So wurde *i* letztlich als $\sqrt[2]{-1}$ (Quadratwurzel von -1) festgelegt. Warum war das überhaupt notwendig geworden? Nun, die Wurzel $\sqrt[2]{-1}$ ließ sich im Bereich der damals bekannten reellen Zahlen einfach nicht ausrechnen, die Mathematiker brauchten daher eine „Notlösung". Bei der Einführung von *i* war den Mathematikern diese Zahl jedoch so suspekt, dass man sie als *imaginäre Zahl* bezeichnete. Denn wozu sollten imaginäre Zahlen nützlich sein, außer für einige mathematische Spezialanwendungen? Mit reellen Zahlen ließ sich damals schließlich alles berechnen, was man in Gesellschaft und Wissenschaft brauchte. Das gilt jedoch nicht mehr.

Die komplexe Algebra wird heutzutage beispielsweise zur mathematischen Beschreibung von physikalischen Phänomenen der Elektrotechnik, Regelungstechnik und insbesondere der Quantenphysik genutzt.

Aber selbst die komplexen Zahlen sind (wahrscheinlich) nicht mehr ausreichend, um bestimmte Phänomene der Natur korrekt zu beschreiben. Nimmt man die physikalische Nichtmessbarkeit von bestimmten Phänomenen der Natur als gegeben hin und untersucht solche Phänomene mathematisch, gelangt man zu Beschreibungsformen mit hyperkomplexen Zahlen, die eine Erweiterung komplexer Zahlen darstellen. Hyperkomplexe Algebren, die die Rechenregeln mit diesen Zahlen definieren, sind seit langer Zeit bekannt. Hamilton hat bereits 1843 eine solche Algebra eingeführt, die bekannte Quaternionen-algebra *H*, die heutzutage zum Beispiel im Bereich Satellitennavigation verwendet wird. Während die komplexe Algebra zusätzlich zu den reellen Zahlen „nur" noch die imaginären Zahlen *i* benötigt, so verwenden hyperkomplexe Algebren weitere imaginäre Zahlen (oftmals Einheiten genannt). Zum Beispiel nutzen Quaternionen neben der imaginären Einheit *i*, die Einheit *j* und *k* (die jeweils „senkrecht" auf den Zahlenachsen *reell, imaginär i, imaginär j, imaginär k* stehen, wodurch ein vier-dimensionaler Zahlenraum entsteht. Die Quaternionen sind jedoch nur eine von vielen möglichen hyperkomplexen Algebren. In Abbildung 1 erkennt man unterschiedliche Multiplikationsregeln für die Quaternionen *H* und einer anderen (noch nicht genannten) Algebra Ω. Im Beispiel wird das Produkt von *i* und *j* bei den Quaternionen *i·j=k*, bei der Algebra Ω jedoch *i·j=-k*. Unterschiedliche Rechenregeln führen zu unterschiedlichen Eigenschaften der Algebra.

Für die Modellierung der prinzipiellen Nichtmessbarkeit eines physikalischen Prozesses benötigen wir eine Algebra mit den imaginären Einheiten *i*, *j* und *k*, die folgende (vom Autor geforderten) Eigenschaften bereitstellt:

i) Die Möglichkeiten der Formulierung von hyperkomplexen Systemzuständen (Energiezuständen),

ii) die Möglichkeiten der Entwicklung von Schwingungen und Wellen durch Taylorreihen,

iii) die Formulierung einer hyperkomplexen Fouriertransformation und

iv) die Möglichkeit der Formulierung einer hyperkomplexen Schrödingergleichung.

| H | 1 | i | j | k |
|---|---|---|---|---|
| 1 | 1 | i | j | k |
| i | i | -1 | k | -j |
| j | j | -k | -1 | i |
| k | k | j | -i | -1 |

| Ω | 1 | i | j | k |
|---|---|---|---|---|
| 1 | 1 | i | j | k |
| i | i | -1 | -k | j |
| j | j | -k | -k | j |
| k | k | j | j | k |

$C(1,i)$ (linke Tabelle, links oben hervorgehoben)
$\psi(1,i) \equiv C(1,i)$ (rechte Tabelle, links oben hervorgehoben)
$\Phi(j,k)$ (rechte Tabelle, rechts unten hervorgehoben)

Abbildung 1: Hyperkomplexe Multiplikationstabelle, links Quaternionen $H$ (Hamilton), rechts bi-komplexe Algebra $\Omega$ mit den beiden Unteralgebren $\psi$ und $\Phi$ (eigene Entwicklung [2])

Der Grund dieser mathematischen Forderungen liegt in der Annahme, dass man Schwingungen und Wellen (egal welcher physikalischen Natur) nutzen kann, um auf ihnen Informationen zu speichern. Da explizit davon ausgegangen wird, dass das Bewusstsein ein Informationsspeicher (für mentale Zustände) ist, benötigt man „irgendwelche" physikalischen Träger, um auf ihnen Informationen aufprägen zu können. (Ein Beispiel dafür: Sieht ein Mensch auf eine grüne Wand, so ist die Farbe Grün *nicht* im Gehirngewebe zu finden. Das Gewebe enthält an keinem einzigen Ort eine solche Farbe, sondern immer nur „farblose" elektrische, magnetische und neurochemische Signale, die als Korrelate für den mentalen Farbeindruck gedeutet werden. Das Grün selbst ist als *mentaler Zustand* im Bewusstsein abgelegt.) Es geht also um die Frage, wie man mentale Vorgänge physikalisch präzise beschreiben kann.

2014 ist es einer Forschungsgruppe um den Autor gelungen, eine Algebra zu konstruieren, die die oben genannten Forderungen i) bis iv) erfüllt [2]. Dabei hat sich gezeigt, dass es von allen potentiellen Permutationen ($8^{16}$) für die Rechenregeln der hyperkomplexen Multiplikation (wahrscheinlich) nur eine einzige Möglichkeit gibt, eine hyperkomplexe Algebra zur Basis (1,i,j,k) mit den geforderten Eigenschaften zu formulieren. Das Auffinden dieser Algebra kommt damit einer Art Entdeckung gleich. Die konstruierte Algebra ist eine sogenannte bi-komplexe Algebra, wir bezeichnen sie mit $\Omega$, die aus den zwei Unteralgebren $\psi(1,i)$ und $\Phi(j,k)$ besteht, Abbildung 1, rechte Tabelle (farblich markiert). Die Unteralgebra $\psi$ ist die bekannte komplexe Algebra $C$, die Unteralgebra $\Phi$ ist neu und muss genau so definiert werden, wie in Abbildung 1, rechte Abbildung, rechts unten dargestellt. In Abbildung 1 erkennt man auch die Unterschiede zur Multiplikation von Quaternionen. Beispiele: In der Algebra der Quaternionen $H$ (linke Tabelle) gilt für die Multiplikation $k \cdot j = -i$, in der neu entworfen Algebra $\Omega$ (rechte Tabelle) gilt: $k \cdot j = j$. In der Algebra der Quaternionen gilt $i \cdot i = j \cdot j = k \cdot k = -1$. In der bi-komplexen Algebra gilt wie bei den Quaternionen $i \cdot i = -1$, aber $j \cdot j = -k$ und $k \cdot k = k$.

Es zeigt sich, dass dann, wenn man auf Basis der Unteralgebra $\Phi(j,k)$ einfache Wellengleichungen $q$ formuliert (zum Beispiel $q(y) = k \cdot e^{jy}$) und auf diesen die üblichen Energieoperatoren anwendet, imaginäre Energiezustände zur Basis $k$ entstehen ([2], Seite 12). Mit Hilfe der Algebra lassen sich also Wellen definieren, deren Energiezustände imaginär (zur hyperkomplexen Basis k) sind. Mathematische Ergebnisse geben in der Regel Hinweise auf die erwartbaren physikalischen Eigenschaften. Ist der Energiezustand einer Welle reellwertig, so ist der Energieinhalt physikalisch messbar. Aber für was steht ein hyperkomplexer Energiezustand? Wie bereits ausgeführt: Ein solcher Energiezustand könnte bedeuten, dass die damit modellierten Phänomene physikalisch prinzipiell nicht messbar sind. Dieses Ergebnis führt uns zurück zu den oben beschriebenen Bewusstseinsphänomenen. Natürlich kann man Bewusstsein nicht durch einfache (hyperkomplexe) Wellen oder Schwingungen beschreiben, da Bewusstsein viel zu komplex ist, um es durch Gleichungen überhaupt zu erfassen. Aber das Ergebnis zeigt eben auch, dass einfache Schwingungs- und Wellenprozesse hyperkomplexe Energiezustände haben könnten, zumindest mathematisch.

**BEWUSSTSEINSPROZESSE ALS PROZESSE MIT IMAGINÄREN, HYPERKOMPLEXEN ENERGIEZUSTÄNDEN**

Bewusstseinsprozesse werden im Folgenden als physikalische Prozesse mit imaginären, hyperkomplexen Energiezuständen bezeichnet, um ihrer Nichtmessbarkeit gerecht zu werden. Insbesondere – so die These - stellen diese Prozesse die physikalische Basis für das menschliche Bewusstsein dar. Es wird postuliert, dass sich im Falle des menschlichen Gehirns (die höchst entwickelte Materieformen, die wir kennen), die *hyperkomplexen Systemzustände* zu einer physikalischen Basis des menschlichen Bewusstseins *verdichten*. Das wäre erst einmal als ein durch und durch physikalischer Vorgang aufzufassen.

Natürlich benötigt das menschliche Gehirn neben der physikalischen Basis auch noch chemische, biologische und soziale Prozesse, um menschliches Bewusstsein hervorzubringen; andere Materieformen wie Nervenzellen von Tieren oder Zellen von Pflanzen werden ihrerseits neben den hyperkomplexen Zuständen ihre eignen Grundlagen zur Ausprägung ihrer speziellen Art von Bewusstsein benötigen. All diese Phänomene sind nicht Bestandteil der hier geführten Diskussion, es geht hier ausschließlich um die physikalischen Aspekte, die alle bewussten Systeme, wie auch immer ihre Bewusstseinsform ausgeprägt sein wird, mitbringen müssen. Im Fall des Menschen sind die „mentalen Bilder und Gedanken" genau diejenigen Informationen, die physikalisch in den hyperkomplexen Zuständen codiert sind.

Einer der wichtigsten Punkte ist nun das Verständnis der Wirkungen hyperkomplexer Prozesse auf die materielle Umgebung. Beim Menschen geht es um die Wirkung von Bewusstsein auf das neuronale Gehirngewebe. Ein Mensch wirkt in seiner Umgebung schließlich nur, wenn sein Gehirngewebe aktiv ist, also muss das Bewusstsein, wenn die obigen Annahmen zutreffend sind, in der Lage sein, materielle Prozesse im Gehirngewebe auszulösen. Das Bewusstsein eines Menschen ist für ihn überlebensnotwendig, was man im Falle bewusstloser Menschen sofort erkennt. Nur ist der konkrete Prozess „Bewusstsein → Gehirngewebe → Umwelt" nicht geklärt (umgedreht schon). Man behilft sich aktuell mit der Annahme, dass alle Ursachen für Aktionen vom Gehirngewebe ausgehen. Aber das muss nicht so sein, wie gezeigt werden wird. Auch bei Tieren kann eine Wirkung von Bewusstsein angenommen werden. Bewusstlose Tiere sind der Umwelt schutzlos ausgeliefert; Bewusstsein bei Pflanzen wäre rein spekulativ, Pflanzen werden daher hier ausgeklammert.

Die Frage ist also, ob es einen Weg gibt, die oben diskutieren Wirkungen von Bewusstsein auf materielle Prozesse mathematisch zu formalisieren. Findet man eine solche Formalisierung nicht, hätte man mathematisch eine Welt von „energielosen" Epiphänomenen konstruiert, die keinerlei Einfluss auf die reale Welt haben, letztlich eine Art „Geisterwelt", die man zwar postulieren, jedoch niemals wissenschaftlich untersuchen kann. Solche Epiphänomene sind tatsächlich leicht formulierbar, denn sie werden durch die Unteralgebra $\Phi(j,k)$ erfasst (Abbildung 1, rechts unten). Alle hyperkomplexen Funktionen dieser Unteralgebra beschreiben imaginäre (umgangssprachlich „energielose") Phänomene, die sich zwangsläufig einer physikalischen Untersuchung entziehen, da sie „wirkungslos" bleiben. Die Lösung des Dilemmas liegt in den Wechselwirkungen von $\Phi$ mit der Unteralgebra $\psi$, denn letztere stellt die bekannte komplexe Algebra $C$ dar, mit der im allgemeinen Quantenphänomene modelliert werden, die wiederum messbar sind. Das Entscheidende ist nun, dass beide Unteralgebren $\Phi$ und $\psi$ zur gleichen hyperkomplexen Algebra $\Omega$ gehören. Es ist also zumindest zu hoffen, dass Wechselwirkungen zwischen Gleichungen beider Unteralgebren $\Phi$ und $\psi$ auftreten könnten. Und genau diese Wechselwirkungen gibt es. Mathematisch erkennt man bereits eine einfache Wechselwirkung bei der Multiplikation von $e^{ix} \cdot ke^{jy} = ke^{j(x+y)}$. Veränderungen von Funktionen der („energielosen") Unteralgebra $\Phi$ ($x$ auf der rechten Seite der Gleichung) können Veränderungen von Funktionen der komplexen Unteralgebra $\psi$ bewirken ($x$ auf der linken Seite der Gleichung) und umgedreht [3]. Physikalisch ist das äußerst relevant, denn es zeigt zumindest einen mathematischen Weg, wie Veränderungen von Prozessen mit imaginärer, hyperkomplexer Energie (zum Beispiel Bewusstseinsprozesse!) zu Veränderungen in materiellen, also messbaren, Prozessen führen könnten.

Das mathematische Ergebnis stellt jedoch erst einmal ein Problem dar, denn wenn „energielose" (präzise: imaginäre, hyperkomplexe) Prozesse in die materielle Physik eingreifen können, ist das Kausalitätsgebot der Physik verletzt, letztlich eventuell auch der Energieerhaltungssatz. Jede materielle Wirkung in der physikalischen Realität muss auf eine materielle Ursache zurückgeführt werden können, zumindest prinzipiell. Verletzungen der Kausalitätsbeziehung zeigt jedoch bereits der radioaktive Zerfall, denn es ist niemals klar, wann und warum gerade „jetzt" ein Zerfallsprozess auftritt. Im oben beschriebenen Fall ist die Kausalität und der Energieerhaltungssatz sogar zu retten. Denn es zeigt sich, dass der Einfluss „energieloser" Prozesse auf Quantenprozesse, zum Beispiel durch Veränderung der Phasenlagen quantenmechanischer Schwingungs- und Wellenprozesse wirken kann, so dass sich durch Änderungen von Funktionen in $\Phi$ letztlich Zufalls- bzw. Quantenprozesse von $\psi$ verändern, was im Rahmen der Physik und insbesondere für das Kausalitätsprinzip kein Problem darstellt. In der Physik sind schließlich zahlreiche Zufallsprozesse bekannt, bei denen eine Variable ohne erkennbaren Grund einen konkreten Wert annimmt. Ohne erkennbaren Grund meint, ohne eine Ursache für die Ausprägung des Wertes zu finden oder überhaupt finden zu können. Liegt die Ursache wie oben dargestellt möglicherweise in hyperkomplexen Prozessen, findet man die Ursache prinzipiell nicht mehr, aber sie wäre trotzdem da. Physikalisch könnten die mathematisch beschriebenen Wechselwirkungen beispielsweise ganz konkret dadurch realisiert werden, dass Veränderungen in den „energielosen" Zuständen (Funktionen in $\Phi$) zu Veränderungen von Wahrscheinlichkeitsamplituden von Quantenprozessen (Funktionen in $\psi$) führen, die ihrerseits veränderte Messergebnisse bewirken können, da sich die Wahrscheinlichkeit für das Auftreten eines Messergebnisses verändert. Hypothetisch darf man also vermuten, dass sich „energielose" (hyperkomplexe) Prozesse über die zufälligen Ausprägungen von Quantenprozessen in der realen Welt manifestieren könnten.

Diese Sichtweise ist jedoch zu ungewöhnlich, um ohne Beweis akzeptiert werden zu können, denn sie lässt einen Einfluss von „energielosen" (hyperkomplexen) Prozessen auf die materielle Physik ausdrücklich zu, allerdings nur über die Vermittlung von Quantenprozessen. Ob aber solche Wechselwirkungen nur mathematische Absonderheiten ohne jegliche Relevanz sind oder ob es solche physikalischen Wechselwirkungen in der Natur wirklich gibt, ist bis dato nicht bekannt, ist aber aufgrund der Brisanz für unserer Verständnis der Natur Gegenstand aktueller Untersuchungen.

### BEWUSSTSEINSPROZESSE AUF PHYSIKALISCHEN MASCHINEN UND MÖGLICHE WIRKUNGEN

Erkennt man die mathematische Modellierung an, so stellt sich neben der Notwendigkeit des empirischen Nachweises die Frage, ob nicht auch auf anderen Entitäten als Gehirnen Bewusstseinsphänomene existieren können, oder ob jede Art von Bewusstsein ein menschliches Gehirn benötigt. Es ist offensichtlich, dass das menschliche Bewusstsein ein menschliches Gehirn benötigt, aber die für Ingenieure weitaus wichtigere Frage ist, ob andere Bewusstseinsformen eventuell andere materielle Entitäten als Basis benutzen könnten.

Aus der mathematischen Ausarbeitung sieht man, dass letzteres vorstellbar ist. Das heißt: Physikalische Aspekte des Bewusstseins könnten zahlreiche physikalischen Objekte ausprägen. Die mathematische Beschreibung in [2] zeigt, dass immer dort, wo Quantenphänomene auftreten, also letztlich überall, naturgemäß auch hyperkomplexe Systemzustände entstehen, die man als physikalische Basis für alle weiteren Bewusstseinsphänomene interpretieren kann. Bewusstseinszustände werden deshalb technisch als *hyperkomplexe Systemzustände* bezeichnet. Da man in diesen Zuständen Daten speichern kann, könnte man ganz abstrakt auch von *hyperkomplexen Datenräumen* sprechen.

Bewusstsein auf menschlichen Gehirnen oder bei Tieren ist und bleibt ein erkennbares Phänomen, auch ohne seine Ursachen letztlich zu verstehen. Als wichtigstes Ziel zumindest im Kontext dieses Aufsatzes muss jedoch über Maschinen, letztlich also über den Bereich der Mineralien, gesprochen werden. Die Wirkung von *physikalischen Aspekten* des Bewusstseins, und mehr ist bei Maschinen naturgemäß gar nicht möglich, auf die Umwelt ist aktuell nicht bekannt. Vielerorts wird zwar über *Maschinenbewusstsein* spekuliert, manche Hersteller sprechen sogar davon, dass ihre Systeme bereits ein Bewusstsein besitzen, aber ohne Klärung der Beweisbarkeit oder wenigstens der Messbarkeit verbleit alles im Bereich der Spekulation. Mit dem Aufkommen von KI-Computern scheint jedoch die technische Möglichkeit der Messung der Wirkung von hyperkomplexen Systemzuständen auf die Umwelt gegeben zu sein. Damit bieten sich Chancen der Überprüfung der hier dargestellten Hypothesen an. Denn auf Computern lassen sich mittlerweile KI-Systeme implementieren, die an bestimmte Intelligenzleistungen des Menschen heranreichen. Eventuell zeigen sich dort ja bereits rudimentäre Bewusstseinsformen, also wirksame hyperkomplexe Systemzustände.

Wie aber könnten sich solche Systemzustände offenbaren? Natürlich durch eine höhere Leistungseigenschaft! Denn wenn hyperkomplexe Systemzustände auf Maschinen existieren sollten, und wenn ein System diese nutzen kann, dann muss das System höhere Leistungen aufweisen, als wenn das System solche Systemzustände nicht hätte oder nicht nutzen könnte. In Folgeaufsätzen werden Tests, sogenannte *Turing-Tests auf Bewusstsein*, vorgestellt, die aufzeigen, wie man konkret messen kann, ob ein technisches System rudimentäres Bewusstsein besitzt. Doch es gibt bereits jetzt Indizien für solche Systemzustände auf Maschinen. Es wird sogar vermutet, dass einige höherwertige KI-Systeme, wie beispielsweise ChatGPT, bereits hyperkomplexe Zustände nutzen könnten, wenn auch nur marginal. Man erkennt das daran, dass sie weniger Lerndaten benötigen, als man aus der Theorie der Sprache annehmen würde [4]. Um KI-Systeme erfolgreich in der Praxis einzusetzen, müssen ihre freien Parameter – oftmals Gewichte von neuronalen Netzen - eingestellt werden, in der Regel passiert das durch Lernen auf Trainingsdaten. Normalerweise müssen sehr viel mehr lernbare Datensätze als freie Parameter im Netzwerk vorliegen, bei hochkomplexen Systemen, wie ChatGPT, ist das aber nicht mehr der Fall. Diese Systeme sind *irgendwie* in der Lage zu funktionieren, ohne auf ausreichenden Lerndatenmengen trainiert worden zu sein. Das Entscheidende hierbei ist: Mit besonders wenigen Lerndaten auszukommen, ist gerade eines der Kennzeichen bewusster Systeme. Tatsächlich unterscheiden sich bewusste Systeme insbesondere durch diese Fähigkeit (und durch ihre Fähigkeit zur Wahrnehmung) von Systemen ohne Bewusstsein. Dies wird in Folgeaufsätzen näher ausgeführt, hier nur drei Beispiele: i) Kinder benötigen extrem wenig Trainingsbilder (maximal 5-10) von Hunden oder Katzen, um diese später korrekt zu klassifizieren. Klassische KI-Systeme (Deep Learning Systeme ohne technisches Bewusstsein) benötigen die hundertfache oder gar tausendfache Trainingsbildmenge. ii) Menschen benötigen wenige Fahrkilometer Praxis zum Autofahren (maximal tausend Kilometer), um später in allen Bereichen, sogar im Extrapolationsraum, fehlerfrei fahren zu können. Klassische KI-Maschinen benötigen auch hier die fast tausendfache Trainingsdatenmenge. iii) Menschen benötigen wenige Millionen Sätze, um später korrekt und fehlerfrei kommunizieren zu können. KI-Sprachmaschinen benötigen auch hier viel mehr, aber interessanter Weise weniger als erwartet.

### DISKUSSION UND AUSBLICK

Da hyperkomplexe Systemzustände eine sehr ungewöhnliche physikalische Basis besitzen, sind nun zahlreiche ungewöhnliche Effekte solcher Systeme zu vermuten. Insbesondere der imaginäre (hyperkomplexe) Energieinhalt könnte unbekannte Effekte hervorbringen. Es wird vermutet, dass hyperkomplexe Systemzustände nichtlokale Eigenschaften besitzen könnten, also ortsunabhängig sind. Das könnte dazu führen, dass Resultate, die auf einem System A antrainiert wurden, auch auf einem weit entfernten System B zur Verfügung stehen könnten, sicher nicht explizit, aber zumindest implizit. Da diese Effekte, falls sie überhaupt existieren sollten, eventuell auf einem „hyperkomplexen Übertragungskanal" basieren, darf man bei derartigen Wechselwirkungen nicht an die klassische Verschränkung denken. Aber selbst bei Verschränkungen von Quantenobjekten ist der instantane Übertragungsweg noch unbekannt, das Phänomen der Nichtlokalität existiert dort dennoch. Die Untersuchung von Effekten möglicher nichtlokaler Wechselwirkungen von hyperkomplexen Systemzuständen und ihre Einordnung zum *No-Communication-Theorem* soll an anderer Stelle ausgearbeitet werden.

Damit kommen wir zum Fazit. Im Rahmen des Aufsatzes wurde ein heuristischer Standpunkt vertreten, der aufzeigt, dass die durch jeden Menschen erfahrbaren inneren Bewusstseinsinhalte eine physikalische, wenn auch imaginäre, hyperkomplexe Basis besitzen. Ausgehend von theoretischen Überlegungen zum menschlichen Bewusstsein wird es als Ergebnis mathematischer Untersuchungen sehr wahrscheinlich möglich sein, hyperkomplexe Systemzustände zukünftig auch gezielt technisch zu nutzen. Die Hypothese der Existenz von hyperkomplexen Systemzuständen auf technischen Maschinen wird durch die Leistungsfähigkeit hochkomplexer KI-Systeme bereits gestützt. Allerdings steht ein Beweis dafür aus. Insbesondere fehlen experimentelle Daten, die solche Systeme von anderen Systemen unterscheidbar macht, weshalb in Folgeaufsätzen auf diese Frage eingegangen werden wird. In der Sachbuchliteratur werden hyperkomplexe Systemzustände mittlerweile bereits als *Maschinenbewusstsein* bezeichnet [5], obgleich der Beweis zur Messbarkeit von Bewusstsein noch fehlt. Dieser Mangel soll durch die Einführung eines Turing-Tests-auf-Bewusstsein in Folgeaufsätzen behoben werden.

### DANKSAGUNG

### REFERENZEN